# Language Identification of Hindi-English tweets using code-mixed BERT

Mohd Zeeshan Ansari
*Department of Computer Engineering*
*Jamia Millia Islamia*
New Delhi, India
mzansari@jmi.ac.in

M M Sufyan Beg
*Department of Computer Engineering*
*Aligarh Muslim University*
Aligarh, India
mmsbeg@hotmil.com

Tanvir Ahmad
*Department of Computer Engineering*
*Jamia Millia Islamia*
New Delhi, India
tahmad2@jmi.ac.in

Mohd Jazib Khan
*Department of Computer Engineering*
*Jamia Millia Islamia*
New Delhi, India

Ghazali Wasim
*Department of Computer Engineering*
*Jamia Millia Islamia*
New Delhi, India

*Abstract*— Language identification of social media text has been an interesting problem of study in recent years. Social media messages are predominantly in code mixed in non-English speaking states. Prior knowledge by pre-training contextual embeddings have shown state of the art results for a range of downstream tasks. Recently, models such as BERT have shown that using a large amount of unlabeled data, the pre-trained language models are even more beneficial for learning common language representations. Extensive experiments exploiting transfer learning and fine-tuning BERT models to identify language on Twitter are presented in this paper. The work utilizes a data collection of Hindi-English-Urdu code-mixed text for language pre-training and Hindi-English code-mixed for subsequent word-level language classification. The results show that the representations pre-trained over code-mixed data produce better results by their monolingual counterpart.

## I. Introduction

The presence of linguistic units from one language in a speech or text from another language is characterized as code mixing. Not only is it utilized in frequently spoken multilingual settings, but it is also employed on social networking platforms in the form of comments and responses, posts, and most notably in chat discussions. The contextualized word embedding trained on huge amounts of text data have recently shown state-of-the-art performance in a variety of natural language processing applications. Models like as BERT and its multilingual variant, m-BERT, depend on huge quantities of unlabeled monolingual text data to construct monolingual and multilingual models that may be utilized for downstream tasks requiring little knowledge of the target language [1]. Several studies in language representation models, such as BERT, have shown the critical relevance of prior knowledge for a variety of downstream tasks, including sentiment analysis, natural language understanding, and neural machine translation. Cross-lingual language models, which build language models from corpora from several languages, have also shown their capacity to achieve state-of-the-art performance in natural language understanding and neural translation [2]. However, Translation Language Models need a costly corpus of parallel sentences for training, and such a dataset may not be accessible in coding mixed languages, as is the case with English-Hindi social media content, where all the words are in roman character. Moreover, some texts may have relatively little code mixing, while others may have frequent and complicated code mixing. In code mixed text of similar languages, the word order of both languages is preserved such as English-Spanish, while extremely dissimilar language pairings adopt the word order of one of the languages, such as Hindi-English pair.

Several natural language processing tasks, such as text classification, named entity recognition, language identification, and others, have been successfully implemented utilizing transformer models [3-6]. Because the majority of the pre-trained transformer models were trained on English data, the majority of the tasks were centered on English language. Despite the fact that there are multilingual models, such as m-BERT [1], the problem arises regarding its ability to represent code mixed data. The m-BERT models do not perform better on Hindi-English code mixed data primarily, they both maintain their separate unique word order, and, secondly, Romanized Hindi does not adhere to conventional lexical norms.

The basic concept in this paper is that we use the transformer to train a language model on large corpus of closely similar languages in code mixed settings, such as Romanized Hindi and Urdu, and then conduct transfer learning architecture on the language with less resources, such as Hindi. Such large mix language corpus is used to train BERT with the masked language model task and the next sentence prediction task. BERT utilizes a masked language model to predict words that are randomly masked or substituted, unlike earlier bidirectional language models that relied on a mixture of two unidirectional language models. Further, BERT is the fine-tuned over the downstream language identification task with Hindi-English code mixed text of relatively small in size. This work examines the BERT for language identification of code-mixed text in a variety of settings including monolingual language model pretraining in English. Moreover, it involves pre-training the BERT model using large Hindi-Urdu-English code-mixed data, then conducting experiments on fine-tuning with RoBERTa for downstream language detection. The findings are compared to the pre-trained BERT baselines.

The overall organization of paper is as follows. The section II presents the recent related work. Section III

discusses the methods of data preparation for the experiments. Section IV presents the various stages involved in code-mixed text language identification using BERT language and classification models. The experimental settings and results are discussed in section V followed by conclusion in section VI.

## II. RELATED WORK

Code-mixed text is a prevalent trend in social media due to which research are being conducted on the extraction of information from such text. The method of POS Tagging was used in code-mixed social media content for several Indian languages [7]. A collaborative effort was formed to extract entities from code-mixed Tamil-English and Hindi-English social media content [8]. Embedding models were used to extract entities from code-mixed Hindi-English and Tamil-English datasets [9]. Numerous natural language processing systems rely on word embedding as an efficient input representations for several tasks such as entity extraction, sentiment analysis, question answering, neural machine translation, and other similar tasks. Gella et al. (2014) develop a language identification method for 28 languages using a synthetically generated code-mixed dataset [10]. Along with the more conventional method of machine learning, some researchers have utilized models based on neural architectures. Chang and Lin (2014) use a recurrent neural network architecture with pre-trained word2vec embeddings to analyze the English-Spanish and Nepali-English datasets from the First Shared Task on Language Identification in Code-Switched Data [11]. Samih et al. (2016) develop a neural network architecture based on LSTMs for the English-Spanish datasets from the Second Joint Task on Language Identification in Code-Switched Data. Their model is composed of word and character representations that have been initialized using pre-trained word2vec embeddings [12].

As a critical component of contemporary natural language processing systems, specialized pre-trained word embeddings may provide substantial gains over embeddings learnt from scratch [13,14]. Additionally, downstream models use generalizations of word embeddings, such as phrase embeddings or paragraph embeddings [15,16]. In earlier works BERT has been adapted to a variety of domains, particularly those with their own complex vocabulary, such as the biomedical domain, legal domain, and financial document processing and scientific texts or publications [17-20]. The majority of these efforts make extensive use of sophisticated methods for mining huge amounts of domain-specific text from the internet, and therefore prefer to train the BERT model from scratch rather than fine-tune the existing BERT checkpoints. While the majority of studies examined domain adaptation by simply continuing training with masked language model goals, several studies examined alternative methods for improving downstream task performance. Ma et al. (2019) achieve domain adaptation via the application of curriculum learning and domain-discriminative data selection [21]. Ye et al. (2020) investigate the adaptability of BERT across languages [22]. Domain adaptation, on the other hand, is not always successful and may result in suboptimal performance. This is dependent on a number of variables, including the degree to which the domains are distinct and the amount of data available [23,24].

## III. DATA PREPARATION

The fact that two data sets in three distinct languages are code-mixed is a characteristic that we exploit, moreover, Hindi and Urdu share large portion of common vocabulary. The term code-mixed refers to sequence of *words* in the same phrase but found in many different languages, the script may remain same or belong to the language. The data we utilize for this assignment are mostly in English texts mixed with Romanized Hindi and Urdu.

### A. Pre-training Dataset

For model pertaining, a composite corpus is constructed by merging the Hindi-English and Urdu-English code mixed datasets [25]. The Hindi-English dataset contains more than 20,000 sentences and Urdu-English dataset contains 19664 sentences which are code-mixed in nature. We exploit the fact that Hindi and Urdu both share a large set of vocabulary with each other.

### B. Fine Tuning Dataset

The dataset is created by finding appropriate Twitter accounts with a significant quantity of Hindi-English code-mixed text and then mining the tweets in those handles. Politics, sports, news, and other topics are carefully targeted in the scrapped tweets. We utilized the tagset of seven entity classes for annotation of tokens, keeping the tweets that included all English and Hindi terms intact: English (En), Hindi (Hi), Universal (Univ), Username (User), Hashtag (Hash), url (U), and Named Entity (NE). The annotation rules are quite similar with those seen in previous Twitter datasets. We observed that the most common tokens, usernames, have highly irregular orthographic patterns. Furthermore, recognizing them would be similar to the Twitter Handle Classification Task, which is not addressed in this paper. Following that, we normalize all of the users, hashtags, and urls by replacing their tokens with their labels, and then we preprocess the dataset. There are many identified entities, which are all labelled as NE. The remaining words are divided into Hi and En tags based on their natural language. Tokens with punctuations, emoticons, and the remainder of the tokens that does not fit into one of the six categories are labelled Univ. Four annotators who can comprehend both Hindi and English carry out the word-level annotation. The data preparation is in line recently adopted guidelines. Table 1 shows the distribution of tags for the dataset prepared by us.

| Table 1. Distribution of train and test tags | | |
|---|---|---|
| *Label* | *# Train words* | *# Test words* |
| English | 283986 | 31554 |
| Hindi | 28269 | 3141 |
| Universal | 15399 | 1711 |
| Hashtags | 1557 | 173 |
| URLs | 110 | 12 |
| Usernames | 873 | 97 |
| Named Entities | 5291 | 587 |

## IV. CODE MIXED BERT MODELS

BERT is a multilayer encoder of the transformer that, unlike previous models for language modelling, is intended to train deep bidirectional representations on unlabeled corpora via joint conditioning in all layers of the model utilizing left and right contexts. BERT receives input in the form of WordPiece [25] or byte level BPE vector representations. We use unsupervised pre-training BERT model, identical to Devlin et al. (2018), followed by supervised fine-tuning on code mixed data, to train the RoBEERTa, adopted from Liu et al [1,26]. The language model is trained on large corpus with BERT and are adapted to the domain by RoBERTa in which they are utilized for word level language identification. This incorporates benefits from domain adaptation by exposing the text in code-mixed settings and enhances its vocabulary using Hindi-English terms. Once the model is completely pre-trained, we use the labelled training and validation data to fine-tune it for the task, which is to predict the language label of each word. The fine tuning involves the BPE tokenization methods.

### A. BERT for Pre-training

BERT accepts as input the concatenation of two segments, which are represented as sequences of tokens by the system. Segments are often made up of more than one phrase or sentence in natural language. These two segments are given to BERT as a sequence of single input, with special tokens of [CLS] and [SEP] denoting the boundaries between them. M and N are the length of both segments which are constrained such that $M + N < T$, where T is a parameter that regulates the maximum length of sequence in the model training. BERT employs the popular transformer architecture of Vaswani et al., 2017 [27].

### B. Subword Vocabulary Generation

Subword vocabularies are made up of several subwords, moreover, they are appropriate balance of word and character vocabularies, recently becoming popular for creating vocabularies. There are primarily two methods for creating subword vocabulary: Byte Pair Encoding (BPE) and WordPiece [28,25]. BPE and WordPiece initialize the vocabulary as a collection of characters and then insert a pair of tokens into the vocabulary repeatedly until the vocabulary size reaches a specified value. Their distinction is in the process of selecting the token pair in each iteration. BPE iteratively substitutes the pair of consecutive characters that appear most often in a text with a single character that does not exist in the text. Additionally, it keeps a mapping database that links each pair of substituted characters to its corresponding character for decoding purposes. Each character is regarded as the most basic element throughout the vocabulary training. This training technique is popularly known as Byte-Level Byte Pair Encoding (BLBPE) because the text is treated as a byte sequence. It is presumed that the algorithm accepts raw text as input, thus the vocabulary building process includes the steps of converting raw text to byte sequence and applying BPE to the byte-level text. The byte-level subwords are useful in situations when the character-level text contains uncommon characters.

### C. RobBERTa for Fine Tuning

For contextualized representations of each word in the input phrase, the model heavily depends on the Transformer-based pre-trained language model RoBERTa [26]. Each phrase is tokenized with the help of byte-pair encoding. We use the AdamW optimizer with RoBERTa, a transformer having a twelve layer architecture with twelve attentional heads and 110M trainable parameters. One distinction from the original BERT model is because RoBERTa provided a new pre-training task, using just the masked language model task rather than the next sentence prediction task. However, during pre-training, it predicts just whether words are masked in certain places inside provided phrases.

## V. EXPERIMENTS AND RESULTS

### A. Baselines

We implement two baselines (1) Pre-trained uncased BERT base for language model training with input representation using WordPiece and training RoBERTa over Hindi-English Twitter dataset; (2) Pre-trained case sensitive BERT base for language pre-training with vocabulary representation of WordPiece and training RoBERTa over Hindi-English Twitter dataset.

### B. Models

Two models are implemented with large dataset for language modelling and Hindi-English Twitter dataset for downstream language identification. The case sensitive BERT language models are trained with Hindi-English-Urdu data of around 40,000 sentences. For each model we utilize separately the subword vocabulary generation models of WordPiece and BLBPE. The RoBERTa is used for the word level classification of tweets in several class labels. The configurations of baselines and code mixed BERT models is presented in Table 1.

### C. Results

This section presents the experimental findings for each of the previously described models. The accuracy, recall, and F1-score of token labels are calculated in the findings. Additionally, the weighted average for all labels is presented. A model is considered to be the best in a setting if its precision, recall, or F1-score is highest. The bold value indicates the maximum value possible for a given configuration.

Table 2. Configuration of baselines and models

| Model Name | Language Model Pretraining | | Subword Vocab | Down Stream Language Identification | |
|---|---|---|---|---|---|
| | BERT variant | Dataset | | BERT variant | Dataset |
| Baseline1 | BERT uncased | pretrained | WP | BERT | Tw[Hi-En] |
| Baseline2 | BERT cased | pretrained | WP | BERT | Tw[Hi-En] |
| Model3 | BERT cased | Hi-En-Ur | WP | RoBERTa | Tw[Hi-En] |
| Model4 | BERT cased | Hi-En-Ur | BLBPE | RoBERTa | Tw[Hi-En] |

Table 3. Performance comparison of models over class labels

| | Precision | | | Recall | | | F-Score | | |
|---|---|---|---|---|---|---|---|---|---|
| | En | Hi | Avr | En | Hi | Avr | En | Hi | Avr |
| Baseline1 | 0.99 | 0.34 | 0.87 | 0.61 | 0.96 | 0.67 | 0.75 | 0.50 | 0.71 |
| Baseline2 | 0.98 | 0.38 | 0.84 | 0.48 | 0.96 | 0.58 | 0.65 | 0.55 | 0.61 |
| Model3 | 0.97 | 0.17 | 0.90 | 0.66 | 0.87 | 0.68 | 0.78 | 0.29 | 0.75 |
| Model4 | 0.97 | 0.37 | 0.90 | 0.82 | 0.89 | 0.82 | 0.89 | 0.52 | **0.84** |
| Veena et al (2017) [29] | NR | NR | NR | NR | NR | NR | 0.65 | 0.82 | 0.80 |
| Shekhar et al (2020) [30] | NR | NR | NR | NR | NR | NR | 0.85 | 0.93 | 0.74 |

| Table 4. Class-wise Performance of Model4 | | | | |
|---|---|---|---|---|
| | Precision | Recall | F-score | Accuracy |
| English | 0.97 | 0.82 | 0.89 | - |
| Hindi | 0.37 | 0.89 | 0.52 | - |
| Universal | 0.56 | 0.82 | 0.66 | - |
| Named Entities | 0.81 | 0.14 | 0.24 | - |
| Overall | 0.90 | 0.82 | 0.84 | 82 |

| Table 5. Performance of Models over Named Entities | | | |
|---|---|---|---|
| | Precision | Recall | F-score |
| Baseline1 | 0.22 | 0.72 | 0.34 |
| Baseline2 | 0.24 | 0.72 | 0.36 |
| Model3 | 0.79 | 0.32 | 0.45 |
| Model4 | 0.81 | 0.14 | 0.24 |

Table 3 shows the performance of the baselines, models and their comparison with recent similar works. It shows that high precision values are obtained for English labels as compared to Hindi labels. However, slightly reverse results are obtained while investigating the recall, that is, high values for Hindi labels and moderately low values for English labels. Observing the overall performance the Model 4 performs best and also outperforms the non-BERT based recent deep neural architecture of Veena et al. (2017) and Shekhar et al. (2020) in terms of weighted average of F-score for all the predicted class labels [29,30]. The Table 4 shows the class-wise performance of Model 4 in which the best results are obtained for English words due to its large distribution in overall data, however, recall of Hindi is best among all. Table 5 presents the performance measures in respect of predicted named entity class in which the baselines show poor precision and good recall, however, on the other hand, the Model3 and Model 4 show opposite results. This is due to the fact that the named entities in the code-mixed text used for model pre-training and fine tuning are in actual Hindi named entities and have morphological characteristics similar to Hindi words. Since, the baselines use the pre-trained BERT language models on English vocabulary, they give low precision and high recall for named entities. Model3 and Model 4 having the language model trained code-mixed data improves the precision on the cost of recall. Fixing this trade-off is left for future work.

## VI. CONCLUSION

BERT based models in different configurations are utilized for pre-training of language models and fine tuning the downstream language identification task. We employ the BERT base models and RoBERTa models for language modelling and classification respectively. We also inspect the effect of vocabulary generation in input representation methods namely, WordPiece and byte level byte pair encodings in classification task and observe that the latter is more effective. We observe that for Hindi-English code mixed language identification both pre-training and fine tuning with code mixed text gives the best F1-score of 0.84 as compared to their monolingual counterparts.